\documentclass[lettersize,journal]{IEEEtran}
\usepackage{booktabs}
\usepackage{multirow}
\usepackage{amsmath,amsfonts}
\usepackage{algorithmic}
\usepackage{algorithm}
\usepackage{array}
\usepackage[caption=false,font=normalsize,labelfont=sf,textfont=sf]{subfig}
\usepackage{textcomp}
\usepackage{stfloats}
\usepackage{url}
\usepackage{verbatim}
\usepackage{graphicx}
\usepackage{cite}
\begin{document}

\title{WildGHand: Learning Anti-Perturbation Gaussian Hand Avatars from Monocular In-the-Wild Videos}

\author{
Hanhui Li$^*$, Xuan Huang$^*$, Wanquan Liu,~\IEEEmembership{Senior Member,~IEEE}, Yuhao Cheng, Long Chen, Yiqiang Yan, Xiaodan Liang,~\IEEEmembership{Senior Member,~IEEE}, Chenqiang Gao$^\dagger$
\thanks{Hanhui Li, Xuan Huang, Wanquan Liu, Xiaodan Liang, Chenqiang Gao are with the School of Intelligent Systems Engineering, Shenzhen Campus of Sun Yat-sen University, P.R. China, 518107. Yuhao Cheng, Long Chen, and Yiqiang Yan are with the Lenovo Research Group, Shenzhen, P.R. China, 518038.}
\thanks{$^*$ Hanhui Li and Xuan Huang contribute equally.}
\thanks{$^\dagger$ Chenqiang Gao is the corresponding author.}
}

\markboth{Journal of \LaTeX\ Class Files,~Vol.~14, No.~8, August~2021}{Li \MakeLowercase{\textit{et al.}}: WildGHand for In-the-Wild Hand Avatars}


\maketitle

\begin{abstract}
Despite recent progress in 3D hand reconstruction from monocular videos, most existing methods rely on data captured in well-controlled environments and therefore degrade in real-world settings with severe perturbations, such as hand-object interactions, extreme poses, illumination changes, and motion blur. To tackle these issues, we introduce WildGHand, an optimization-based framework that enables self-adaptive 3D Gaussian splatting on in-the-wild videos and produces high-fidelity hand avatars. WildGHand incorporates two key components: (i) a dynamic perturbation disentanglement module that explicitly represents perturbations as time-varying biases on 3D Gaussian attributes during optimization, and (ii) a perturbation-aware optimization strategy that generates per-frame anisotropic weighted masks to guide optimization. Together, these components allow the framework to identify and suppress perturbations across both spatial and temporal dimensions. We further curate a dataset of monocular hand videos captured under diverse perturbations to benchmark in-the-wild hand avatar reconstruction. Extensive experiments on this dataset and two public datasets demonstrate that WildGHand achieves state-of-the-art performance and substantially improves over its base model across multiple metrics (e.g., up to a $15.8\%$ relative gain in PSNR and a $23.1\%$ relative reduction in LPIPS). Our implementation and dataset are available at \url{https://github.com/XuanHuang0/WildGHand}.
\end{abstract}

\begin{IEEEkeywords}
Hand avatar, 3D Gaussian splatting, perturbation modeling, in-the-wild video.
\end{IEEEkeywords}

\section{Introduction}
\label{sec:intro}
\IEEEPARstart{H}{ands} play a central role in human interaction in both the physical world and immersive digital experiences. With the rapid development of virtual/augmented reality and embodied intelligence, the demand for realistic, personalized hand avatars is increasing, and recent advances in differentiable rendering \cite{NeRF2020, Kerbl_Kopanas_Leimk_2023} have substantially improved the fidelity of reconstructed avatars. Existing methods \cite{huang20243d, mundra2023livehand, chen2023hand, guo2023handnerf} leverage large-scale data captured in carefully controlled environments (e.g., studios with hundreds of calibrated cameras \cite{moon2020interhand2}) and have demonstrated impressive performance in generating high-quality hand avatars.

However, the high data requirements and limited data diversity of existing approaches severely limit their usability and generalization. To alleviate these issues, several recent studies exploit short monocular videos \cite{moon2024authentic, zheng2024ohta} or even single images \cite{huang20243d, huang2024learning}. To simplify the reconstruction problem, these methods typically assume that both the hands and their surrounding environments remain static. While this assumption reduces the need for large-scale captures, it breaks down in uncontrolled in-the-wild scenarios, where hand observations are often affected by occlusions and extreme poses, or the captured images may further degrade due to motion blur and illumination variations. 

A complementary line of research focuses on dynamic or distraction-aware reconstruction \cite{Martin-Brualla_Radwan_Sajjadi_Barron_Dosovitskiy_Duckworth_2021, Sabour_Vora_Duckworth_Krasin_Fleet_Tagliasacchi_2023, sabour2024spotlesssplats, ungermann2024robust, markin2024t, kulhanek2024wildgaussians}, which leverages geometric, appearance, or semantic consistency across frames to improve robustness. For example, \cite{markin2024t} combines an unsupervised classifier, a segmentation model, and an object tracker to identify transient objects, while \cite{kulhanek2024wildgaussians} leverages DINO features \cite{oquab2024dinov2} with uncertainty-aware optimization to better handle appearance changes. Overall, these methods improve the reconstruction quality of predominantly static scenes in the presence of transient objects, underscoring the importance of explicitly modeling perturbations.

Nevertheless, these dynamic models are not directly applicable to in-the-wild hand avatar reconstruction for two key reasons. First, most of them are designed to handle \textbf{transient distractions}, e.g., moving objects or short-term occluders that can be separated from the target using auxiliary cues such as segmentation or tracking. In contrast, perturbations in hand-centric in-the-wild videos are often \textbf{global} (e.g., illumination changes affecting the entire image) and \textbf{persistent} (lasting for extended time intervals). Therefore, perturbations in our task cannot be treated as sparse, short-lived outliers. Second, these methods are primarily developed for rigid or mildly deformable scenes, whereas hands exhibit highly articulated motions, frequent self-occlusions or hand-object occlusions, and rapid pose changes. These properties make hand avatar reconstruction particularly sensitive to corrupted supervision. As a result, naive optimization can either underfit the true hand appearance or geometry, or overfit to perturbations, leading to an underfitting-overfitting dilemma under real-world perturbations. 

\begin{figure*}[!t]
  \includegraphics[width=\textwidth]{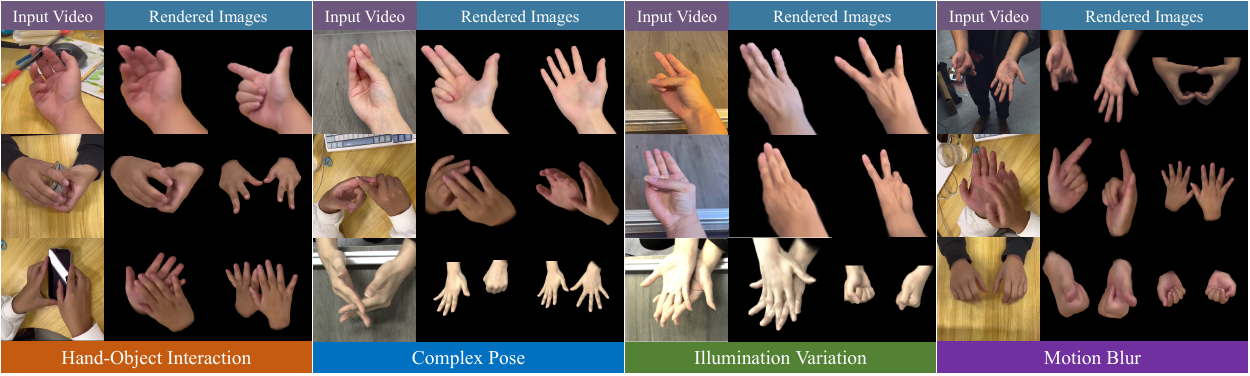}
  \caption{We present WildGHand, a novel Gaussian splatting framework for generating realistic hand avatars from short monocular videos exhibiting challenging perturbations, including hand-object interactions, complex poses, illumination variations, and motion blur.}
  \label{fig:banner}
\end{figure*}

To address the above limitations and enable high-fidelity hand avatar reconstruction from monocular in-the-wild videos under perturbations, we propose a novel framework, dubbed WildGHand. The key idea of WildGHand is to explicitly disentangle perturbations from the underlying hand content during optimization. Concretely, WildGHand extends the optimization-based 3D Gaussian splatting (3DGS) framework \cite{Kerbl_Kopanas_Leimk_2023} with two components: a dynamic perturbation disentanglement (DPD) module and a perturbation-aware optimization (PAO) strategy. The DPD module adopts a lightweight multilayer perceptron to model perturbations as temporally weighted biases of 3D Gaussian attributes. These biases are learned on training frames and removed at inference, thereby reducing the risk of overfitting to perturbations. Meanwhile, the PAO strategy leverages semantic hand structures and reconstruction errors to localize perturbed regions and generate anisotropic weighted masks that provide reliable supervision for model optimization. These two components enable our framework to generate robust and realistic hand renderings under various perturbations, as shown in Figure \ref{fig:banner}.

Moreover, since existing benchmarks are either collected in well-controlled environments \cite{moon2020interhand2,chao2021dexycb,fan2023arctic} or contain only limited perturbation diversity \cite{karunratanakul2023harp,moon2024authentic,pavlakos2024reconstructing,grauman2022ego4d}, they are insufficient for evaluating hand avatar reconstruction under realistic in-the-wild environments. To bridge this gap, we systematically categorize common perturbations and curate a challenging dataset accordingly. As summarized in Table~\ref{tab:dataset_comparison}, our hand with perturbation (HWP) dataset covers four representative challenging factors in both single-hand and interacting-hand scenarios, including hand-object interactions, complex poses, illumination variations, and motion blur. Importantly, each sequence contains multiple perturbation types to reflect real-world capture conditions, while we additionally provide a clean test clip to enable reliable and fair evaluation. The dataset further includes diverse hand gestures and daily interactions, such as shuffling cards, spinning pens, and applying hand cream. Experiments on this dataset and two public benchmarks demonstrate that WildGHand produces high-quality personalized hand avatars from complex in-the-wild captures and achieves state-of-the-art performance.

In summary, the contributions of this paper can be summarized as follows:

$\bullet$ We introduce WildGHand, an optimization-based 3D Gaussian splatting framework for reconstructing high-fidelity hand avatars from short monocular \emph{in-the-wild} videos under severe perturbations.

$\bullet$ We propose a dynamic perturbation disentanglement (DPD) module that models perturbations as temporally weighted attribute biases during optimization and \emph{removes} these biases at inference to mitigate overfitting to corruptions.

$\bullet$ We develop a perturbation-aware optimization (PAO) strategy that generates anisotropic per-frame weighted masks to down-weight unreliable regions, thereby improving robustness to both spatial and temporal perturbations.

$\bullet$ We curate a challenging in-the-wild hand video dataset with diverse perturbations to benchmark robust hand avatar reconstruction.

\section{Related Work}
\label{sec:RelatedWork}

\subsection{Differentiable Rendering}

3D reconstruction via differentiable rendering has attracted extensive attention due to its ability to jointly optimize geometry and appearance from image observations. Representative approaches include neural radiance fields (NeRF) \cite{NeRF2020} and 3D Gaussian splatting (3DGS) \cite{Kerbl_Kopanas_Leimk_2023}, which have demonstrated high-fidelity results in a variety of downstream applications such as novel view synthesis \cite{chen2024pgsr, zheng2024gps, niedermayr2024compressed} and human avatar creation \cite{moon2024expressive, jiang2022neuman, guo2023vid2avatar, jiang2023instantavatar, shen2023x}. From the perspective of object and scene representation, existing differentiable renderers can be broadly categorized into (i) \emph{implicit} methods that parameterize targets as continuous functions (e.g., NeRF and its variants), and (ii) \emph{explicit} methods that represent a target using discrete primitives with learnable attributes (e.g., Gaussian points in 3DGS). Implicit representations are expressive and naturally support continuous, arbitrarily high resolution. However, they are often computationally expensive at rendering time. In contrast, explicit representations, particularly 3DGS, achieve efficient rasterization-based rendering with competitive visual quality, and have thus become increasingly popular in practice.

Specifically, owing to the advantages of function-based representations that provide a unified modeling and optimization paradigm, recent NeRF-based methods have been extended to incorporate additional inductive biases and signals, such as explicit physical models \cite{liui2, li2025lirm}, higher-level relational or functional reasoning \cite{koch2025relationfield, weijleropenhype}, and complementary modalities \cite{cao2025universe}. For instance, \cite{liui2} introduces a general radiative physical formulation that combines emission, absorption, and scattering, which can handle various media like underwater and low-light scenes. \cite{koch2025relationfield} proposes a framework that can infer inter-object relationships from NeRFs. \cite{cao2025universe} leverages scene priors from video diffusion models to improve consistency of reconstructed results. 

As for 3DGS, recent research has focused on improving its representational capacity to better handle complex appearance and motion patterns, such as specular surfaces with strong reflections \cite{tang2025spectre}, articulated objects \cite{guo2025articulatedgs}, parameterized objects \cite{gao2025curve}, and fine-grained human details (e.g., faces and hair) \cite{kim2025haircup}. These extensions are often enabled by introducing stronger priors and additional optimization objectives. In particular, video diffusion priors have attracted growing interest \cite{zhong2025taming, schwarz2025generative, li2025trace, go2025videorfsplat}, as they can encode rich geometry and appearance statistics and encourage cross-view consistency. Beyond video diffusion models, several works incorporate semantic priors from multimodal large language models \cite{zhao2025physsplat, li20254d} or texture priors from image diffusion models \cite{zhang2025gap, ma2026ergo}. While such large-scale priors can improve robustness and generalization under challenging capture conditions, they typically introduce substantial computational and engineering overhead, which may limit their applicability in per-scene optimization pipelines. In contrast, our method aims to improve robustness to in-the-wild perturbations without relying on heavy external generative models. Specifically, we employ a lightweight MLP to model perturbations as attribute-level biases with temporal weights, and combine it with a perturbation-aware optimization strategy to reduce the influence of unreliable regions during training. This design achieves a practical balance between efficiency and effectiveness, making it well-suited for robust hand avatar reconstruction from monocular in-the-wild videos.

\begin{table*}[!t]
    \centering
    \small
    \setlength{\tabcolsep}{4pt}
    \renewcommand{\arraystretch}{1}
    \caption{Detailed comparison between the proposed HWP dataset and existing hand video datasets.}
    \begin{tabular}{lcccccccc}
    \toprule
    Dataset & Env. & Clean Testset & Single Hand & Int. Hand & Hand-Object Int. & Complex Poses & Illum. Var. & Motion Blur \\
    \midrule
    InterHand2.6M~\cite{moon2020interhand2} & Lab & \checkmark & \checkmark & \checkmark &  &  &  &  \\
    HARP~\cite{karunratanakul2023harp} & Wild & \checkmark & \checkmark &  &  &  &  &  \\
    UHM~\cite{moon2024authentic} & Wild & \checkmark & \checkmark &  &  &  &  &  \\
    DexYCB~\cite{chao2021dexycb} & Lab &  & \checkmark &  & \checkmark & \checkmark &  &  \\
    EpicKitchen~\cite{Damen2018EPICKITCHENS} & Wild &  &  & \checkmark & \checkmark & \checkmark & \checkmark & \checkmark \\
    HInt~\cite{pavlakos2024reconstructing} & Wild &  & \checkmark & \checkmark & \checkmark & \checkmark &  & \checkmark \\
    ARCTIC~\cite{fan2023arctic} & Lab &  &  & \checkmark & \checkmark & \checkmark &  &  \\
    Ego4D~\cite{grauman2022ego4d} & Wild &  &  & \checkmark & \checkmark & \checkmark & \checkmark & \checkmark \\\hline
    \textbf{HWP (Ours)} & Wild & \checkmark & \checkmark & \checkmark & \checkmark & \checkmark & \checkmark & \checkmark \\
    \bottomrule
    \end{tabular}
    \label{tab:dataset_comparison}
\end{table*}

\subsection{Dynamic Modeling}

Despite the success of differentiable rendering techniques, most existing methods implicitly assume that the underlying scene or subject is view- and time-consistent, i.e., the observations are not substantially affected by environmental changes such as moving distractors, illumination variations, or motion blur. However, this assumption is often violated in real-world capture conditions. 

To improve robustness, a line of work on NeRFs identifies and treats distractors as outliers and reduces their influence by reweighting or truncating the reconstruction loss \cite{Sabour_Vora_Duckworth_Krasin_Fleet_Tagliasacchi_2023, Martin-Brualla_Radwan_Sajjadi_Barron_Dosovitskiy_Duckworth_2021, chen2024nerf, Wu_Zhong_Tagliasacchi_Cole_Research}. RobustNeRF \cite{Sabour_Vora_Duckworth_Krasin_Fleet_Tagliasacchi_2023}, for example, detects outliers by ranking pixel residuals and applying a blur kernel to exploit their spatial smoothness. NeRF-W \cite{Martin-Brualla_Radwan_Sajjadi_Barron_Dosovitskiy_Duckworth_2021} relaxes the consistency assumption by learning per-image appearance embeddings and decomposing scenes into static and transient components. NeRF-HuGS \cite{chen2024nerf} further combines heuristics with segmentation models to enhance robust hand identification and mitigate interference from hand-irrelevant regions.

More recently, several methods have investigated robust training of 3DGS for wild-captured data \cite{sabour2024spotlesssplats, Kirillov_Mintun_Ravi_Mao_Rolland_Gustafson_Xiao_Whitehead_Berg_Lo_et, wang2024we, xu2024wild, zhang2024gaussian}. Similar to RobustNeRF, SpotlessSplats \cite{sabour2024spotlesssplats} performs unsupervised outlier detection in a learned feature space to identify distractors from rendering errors. Robust-3DGS \cite{ungermann2024robust} integrates Segment Anything \cite{Kirillov_Mintun_Ravi_Mao_Rolland_Gustafson_Xiao_Whitehead_Berg_Lo_et} with a neural classifier to refine segmentation masks. 

Several methods \cite{sun2025splatflow, kwon2025efficient, yao2025riggs, luoinstant4d} make their efforts to extend 3DGS to 4DGS by using time-varying Gaussian attributes or deformation fields. For example, RigGS \cite{yao2025riggs} proposes to animate canonical 3D Gaussians by sparse 3D skeletons with learnable skin deformations and pose-dependent detailed deformations. Except for skeletons, sparse motion controls like Hermite splines \cite{park2025splinegs} have been used to drive 3D Gaussians. MoSca \cite{lei2025mosca} introduces the concept of 4D motion scaffolds, which are structured graphs that Gaussians can be anchored on and manipulated with priors extracted by vision foundation models. 7DGS \cite{gao20257dgs} even includes 3D viewing directions into the 4D Gaussian representation, which show superior performance compared to 4DGS.

Nevertheless, most of these approaches focus on short-lived, transient distractors (e.g., moving objects), whereas in-the-wild hand avatar reconstruction often involves broader and longer-lasting perturbations (e.g., global illumination changes), which remain challenging to handle.

\subsection{3D Hand Avatars}
Numerous works \cite{huang20243d, mundra2023livehand, chen2023hand, guo2023handnerf, iwase2023relightablehands} have achieved impressive results in creating hand avatars from studio-captured datasets. LiveHand \cite{mundra2023livehand}, built on NeRF, enables real-time rendering through low-resolution training and 3D-consistent super-resolution. HandAvatar \cite{chen2023hand} introduces a high-resolution MANO-HD model and separates hand albedo and illumination based on pose. RelightableHands \cite{iwase2023relightablehands} presents a novel relighting approach for hand rendering. 

However, these methods rely on deliberately collected data with expensive cameras, tracking devices, and annotation tools, which limits their real-world applications. To address these challenges, recent research has focused on creating personalized hand avatars from short monocular videos \cite{karunratanakul2023harp, moon2024authentic} or even a single image \cite{huang2024learning, zheng2024ohta, potamias2023handy}. HARP \cite{karunratanakul2023harp} optimizes hand avatars from mobile phone sequences. Besides, instead of neural implicit models, it uses a mesh-based parametric hand model with explicit texture representation for better robustness. UHM \cite{moon2024authentic} offers a universal solution for representing hand meshes across arbitrary identities and adapting to individuals with a phone scan. By integrating tracking and modeling into a single stage, UHM \cite{moon2024authentic} overcomes the error accumulation problem, delivering more accurate results. Creating hand avatars from a single image with limited information requires strong prior knowledge. Handy \cite{potamias2023handy}, trained on a diverse dataset of over 1,200 subjects, uses a GAN-based approach to accurately reconstruct 3D hand shape, pose, and textures. OHTA \cite{zheng2024ohta} separates hand representation into identity-specific albedo and transferable geometry for realistic one-shot reconstruction. InterGaussianHand \cite{huang2024learning} combines learning-based features for cross-subject generalization and identity maps for optimization, with an interaction-aware attention module and Gaussian refinement for improved interacting hand rendering.

Moreover, from the perspective of perturbation modeling, current methods mainly focus on hand-object interactions \cite{pang2025manivideo, fan2025re, wang2025dreamactor}, while other challenging factors like illumination changes remain relatively under-explored. Besides, methods of these type usually leverage image or video diffusion models to synthesize animated hands, rather than optimizing an explicit 3D representation via differentiable rendering. For example, \cite{chen2025foundhand} trains a hand image generation model using approximately 10M images. While such data-driven generative pipelines can produce visually plausible results, they typically incur substantial training costs and may suffer from inconsistency or the Janus issue due to the lack of explicit geometric constraints.

Despite the above significant progress in monocular hand reconstruction, a substantial gap remains between current methodologies and their applicability in real-world environments, as they assume near-ideal capturing conditions. On the other hand, we aim at addressing scenarios where diverse perturbations are present, thereby facilitating more robust and real-life applications.

\section{Methodology}
\label{sec:Methodology}
In this section, we present the details of the proposed WildGHand framework (Sec. \ref{sec: framework}) for reconstructing hand avatars from monocular videos that exhibit challenging factors, such as hand-object obstructions, complex poses, motion blur, and illumination variations. Specifically, WildGHand introduces two core components to improve the robustness of the inverse renderings of hands against perturbations caused by the above factors: (i) a dynamic perturbation disentanglement paradigm that explicitly represents perturbations as temporally weighted biases of 3D Gaussians (Sec. \ref{sec: DPD}); and (ii) a perturbation-aware optimization strategy (Sec. \ref{sec: optimization}) that provides anisotropic per-frame weighted masks for fitting 3D Gaussians to videos. These two components enable our method to achieve a better balance between underfitting dynamic hands and overfitting to perturbations, thereby facilitating the generation of high-fidelity personalized hand avatars.

\begin{figure*}[!t]
  \centering
   \includegraphics[width=1.0\hsize]{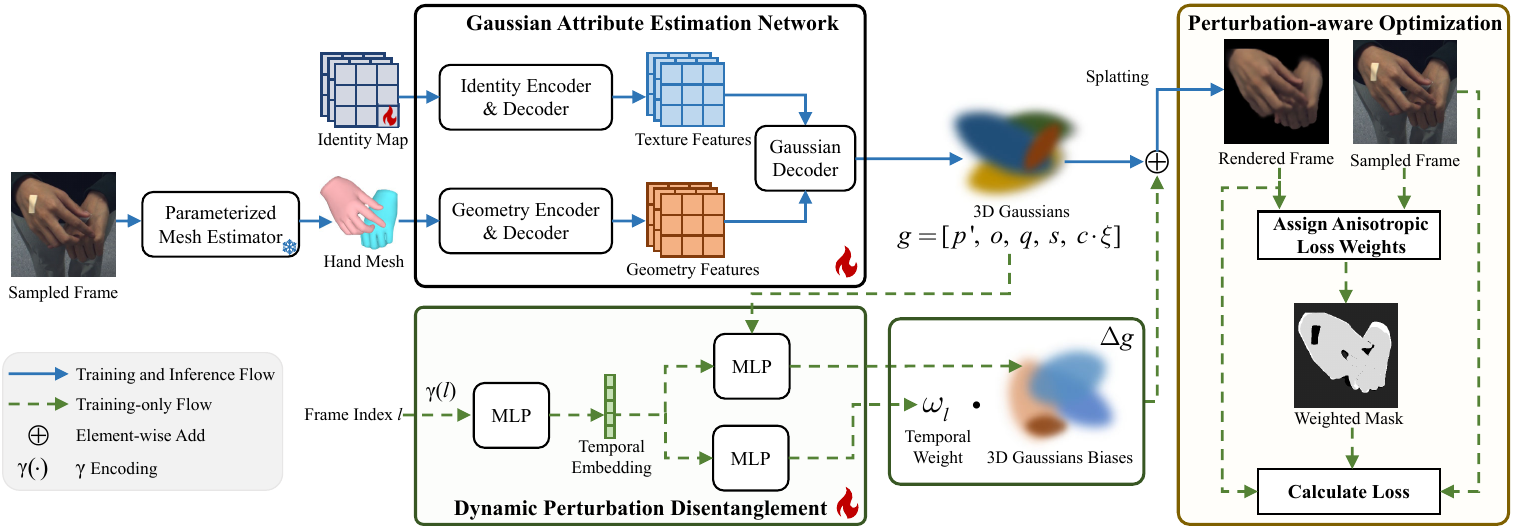}
   \caption{The proposed WildGHand framework. Given a monocular video affected by perturbations, WildGHand introduces two key components to achieve the robust estimation of 3D Gaussians, including a lightweight dynamic perturbation disentanglement (DPD) module and a perturbation-aware optimization (PAO) strategy. The DPD module represents potential perturbations by biases of Gaussian attributes, which are optimized guided by the weighted masks predicted by the PAO strategy. During inference, the optimized biases are removed to render perturbation-free images.}
   \label{fig: overview}
\end{figure*}

\subsection{Preliminary}
We adopt 3D Gaussian splatting (3DGS) \cite{Kerbl_Kopanas_Leimk_2023} to implement our differentiable renderer, which is an efficient rasterization-based rendering method that projects a set of 3D Gaussian spheres onto the image plane. Each Gaussian sphere $g=[p, o, q, s, c]$ is governed by multiple attributes, i.e., a center position $p \in \mathbb{R}^{3}$, an opacity value $o \in \mathbb{R}$, a quaternion $q \in \mathbb{R}^4$ representing rotation, an anisotropic scaling vector $s \in \mathbb{R}^3$, and a color value $c \in \mathbb{R}^3$. Conventional methods \cite{huang2024learning,zhao2024gaussianhand,moon2024expressive} determine the number and the attributes of 3D Gaussians by fitting the input video directly, which are inevitably contaminated by noise and hand-irrelevant information. Therefore, instead of adopting this naive fitting strategy, we propose a self-adaptive optimization paradigm in WildGHand to handle this challenge.

\subsection{Overall Framework}
\label{sec: framework}

\textbf{Task formulation}. 
Given a short monocular hand video, our goal is to reconstruct an animatable hand avatar capable of rendering high-quality images under arbitrary poses and viewpoints. To this end, we adopt the MANO-HD hand model \cite{chen2023hand} that parameterizes the $SE(3)$ transformations of a 3D hand mesh template by a hand pose vector $\theta \in \mathbb{R}^{16 \times 3}$, a hand shape vector $\beta \in \mathbb{R}^{10}$, and a translation vector $t \in \mathbb{R}^{3}$ describing the global position of the hand. Let $h=[\theta, \beta, t]$ denote the concatenation of these parameters. With $h$ and a camera parameter vector $d \in \mathbb{R}^{25}$ describing the intrinsic and extrinsic camera properties \cite{hartley2003multiple}, the hand avatar can be formulated within the differentiable rendering framework, i.e., $\mathcal{R}(h, d) = I$, where $I \in \mathbb{R}^{H \times W \times 3}$ denotes the rendered RGB image with resolution $H \times W$. For clarity, here we assume the given video contains only a single hand; however, our framework can be seamlessly extended to scenarios of interacting hands by independently processing each hand. Unlike existing methods \cite{huang20243d, huang2024learning, moon2024authentic, zheng2024ohta} that are designed for videos captured in clean and static environments, we aim at in-the-wild hand videos with unknown perturbations that hinder the reconstruction process.

The overall framework of WildGHand is shown in Fig.~\ref{fig: overview}, which adopts the per-scene optimization paradigm following \cite{zheng2024ohta,huang2024learning}. Given a training frame $I_l$ sampled from a video comprising $L$ frames, $l \in \{1, \ldots, L\}$, we employ an off-the-shelf estimator \cite{lin2024omnihands} to obtain the hand parameters $h$ and the camera parameter vector $d$. The hand parameters are used to deform the 3D hand mesh template of MANO-HD \cite{chen2023hand} and generate a posed hand mesh. Let $\overline{\mathcal{V}}, \mathcal{V} \in \mathbb{R}^{N\times3}$ denote the $N=49{,}281$ vertices of the template and the posed mesh, respectively. We initialize a set of 3D Gaussians $\mathcal{G}=\{g_n \mid n=1, \ldots, N\}$ centered at $\mathcal{V}$, and estimate their attributes iteratively by jointly optimizing a latent identity map and a Gaussian prediction network. This optimization ensures that the rendered image $\mathcal{R}(h, d)$ approximates $I_l$. The latent identity map $m \in \mathbb{R}^{H\times W\times D}$ has $D$ channels and is used to encode the characteristics of the target hand, while the Gaussian prediction network $f$ is pretrained on the large-scale InterHand2.6M dataset \cite{moon2020interhand2} to leverage cross-subject priors. 

Specifically, $f$ takes as input geometry features extracted from $\mathcal{V}$ and texture features extracted from $m$, and predicts per-Gaussian attributes. $f$ is implemented by a dual-branch encoder--decoder backbone followed by four MLP-based prediction heads. The backbone architecture is identical to that proposed in our previous work \cite{huang2024learning}, which separately extracts geometry features $x_g$ from $\mathcal{V}$ and texture features $x_t$ from $m$. The four prediction heads are designed for distinct purposes: (i) the first head predicts the Gaussian attributes $g$ based on $x_g$ and $x_t$; (ii) the second head predicts a shadow coefficient $\xi \in (0,1)$ from $x_g$ to modulate the color attribute in $g$ as $c \cdot \xi$, which improves the disentanglement of shadow and albedo for rendering; (iii) the third head combines $x_t$ with positional encoding $\gamma$ \cite{NeRF2020} to predict offsets for the template $\overline{\mathcal{V}}$, allowing more flexible deformations to fit various hands; and (iv) the fourth head refines the hand parameters $h$ by estimating the corresponding biases $\Delta h$ conditioned on $x_g$ and $x_t$. Subsequently, we update the center of each 3D Gaussian for the next optimization step as $p' = p_g + v'$, where $p_g$ denotes the center position predicted by the first head and $v'$ denotes the corresponding mesh vertex derived from the predictions of the third and fourth heads.

To model potential perturbations in the input video, we assume that they can be represented as additive biases to the predicted Gaussian attributes. Specifically, the per-Gaussian attributes for a training frame are defined as
\begin{equation}
    \tilde{g} = g + \Delta g,
    \label{eq: perturbation}
\end{equation}
where $\Delta g$ denotes the biases. To estimate $\Delta g$ effectively, we introduce a lightweight module in parallel to the Gaussian prediction network and devise a perturbation-aware optimization strategy to guide its optimization. The details of these two components are elaborated in the subsequent sections. Our design is based on the observation that mesh-based hand representations are relatively stable across frames, while perturbations are abrupt and noisy. Hence, the Gaussian prediction network, with greater modeling capacity, is better suited to capture the canonical component $g$, while the lightweight module focuses on modeling $\Delta g$. Consequently, \emph{we can optimize $\tilde{g}$ during per-scene optimization while employing $g$ for inference}, thereby mitigating the trade-off between overfitting to perturbations and underfitting the underlying hand appearance.

\subsection{Dynamic Perturbation Disentanglement}
\label{sec: DPD}

\begin{figure*}[!h]
  \centering
   \includegraphics[width=1.0\hsize]{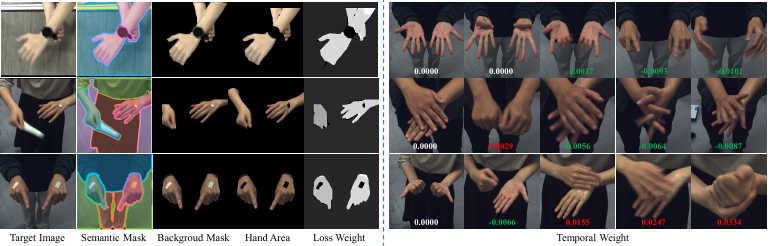}
   \caption{Illustration of the proposed PAO and DPD modules. Left: Our perturbation-aware optimization (PAO) strategy segments the hand regions and leverages reconstruction error to generate weighted masks to guide the optimization of 3DGS. Right: The temporal weights estimated by our dynamic perturbation disentanglement (DPD) module that reflect the strengths of perturbations. Partial perturbations (e.g., occlusions) tends to have smaller weights (labeled in green), while holistic perturbations (e.g., motion blur) have larger weights (labeled in red).}
   \label{fig: loss_weight}
\end{figure*}

Considering that perturbations are time-varying and challenging to model using static representations, we introduce a dynamic perturbation disentanglement (DPD) module that integrates temporal weights into the estimation of Gaussian attribute biases. The DPD module adopts a shallow architecture consisting of three MLP-based blocks, thereby preventing it from overpowering the Gaussian attribute prediction network. 

Formally, inspired by the positional encoding mechanism introduced in \cite{NeRF2020}, the DPD module first computes a high-dimensional positional encoding $\gamma(l)$ of the frame index $l$ to facilitate fine-grained temporal modeling. An MLP-based encoder is then applied to obtain the temporal embedding $z_l$ as follows:
\begin{equation}
    z_l = \phi(\gamma(l)),
\end{equation}
\begin{equation}
\begin{aligned}
\gamma(l)=\bigl[\, & \sin(2^{0}\pi l/L),\,\cos(2^{0}\pi l/L),\,\ldots,\\
                   & \sin(2^{K}\pi l/L),\,\cos(2^{K}\pi l/L)\,\bigr],
\end{aligned}
\end{equation}
where $K$ is the user-specified encoding length, and we set $K=9$ in this paper. The temporal embedding $z_l$ serves two purposes. First, $z_l$ is concatenated with the predicted Gaussian attributes $g_l$ (with the Gaussian/vertex index in $g_l$ omitted for brevity) to predict the biases. Second, $z_l$ is used to predict a global scaling factor $\omega_l \in (-1, 1)$ that controls the influence of the predicted biases. This can be formulated as
\begin{equation}
    \Delta g_l = \omega_l \cdot \varphi([z_l, g_l]), \quad \omega_l = 2\sigma(\psi(z_l)) - 1,
\end{equation}
where $\varphi$ and $\psi$ are each implemented via an MLP, and $\sigma$ denotes the sigmoid function. Furthermore, we randomly set $\omega_l$ to $0$ with a probability of $30\%$ so that the Gaussian attribute prediction network can fit each frame independently, which encourages the network to capture the principal components of the Gaussian attributes.

\noindent \textbf{Discussion}. Compared with existing methods for deformable 3D Gaussians and 4D Gaussians \cite{wu20244d,yang2024deformable}, the role of our temporally weighted Gaussian attribute biases is fundamentally different. Most existing methods utilize biases to facilitate more flexible deformations; that is, larger bias magnitudes are often beneficial. In contrast, our approach uses biases to explicitly model and mitigate perturbations, and therefore favors smaller bias magnitudes. This behavior is supported by Fig.~\ref{fig: loss_weight}, which visualizes the learned temporal weights for frames affected by occlusions and motion blur. We observe that the temporal weights are proportional to perturbation strengths. For example, at the bottom of Fig.~\ref{fig: loss_weight}, the weight of the frame with the most severe perturbations (last column) is approximately five times larger than that of a clean frame (second column). Moreover, the absolute values of these temporal weights remain small (ranging from $0.0066$ to $0.034$), which is reasonable because the temporal weights are used to scale the bias terms rather than govern the main components of the Gaussian attributes.

\subsection{Perturbation-aware Optimization}
\label{sec: optimization}

In addition to the DPD module, which facilitates temporal perturbation modeling, we propose a perturbation-aware optimization (PAO) strategy that spatially identifies intra-frame perturbations, thereby enabling comprehensive suppression across both spatial and temporal dimensions. The core intuition behind PAO is that regions affected by perturbations are inherently more difficult to synthesize; hence, areas with low rendering quality are more likely to be corrupted by perturbations. Therefore, we adaptively reduce the loss weights assigned to these regions during optimization to mitigate their adverse impact.

Specifically, we utilize the Segment Anything Model (SAM) \cite{Kirillov_Mintun_Ravi_Mao_Rolland_Gustafson_Xiao_Whitehead_Berg_Lo_et} to segment the input image $I_l$ into $U$ regions, and we let $\mathcal{Y} = \{y_u \mid u=1, \ldots, U\}$ denote the corresponding binary masks. One advantage of SAM is that it is promptable, which allows us to use the hand root as a prompt to conveniently locate the hand. Let $y_h$ denote the mask of the target hand. We assign an anisotropic loss weight to each region as
\begin{equation}
    \lambda_{u} =\alpha \cdot \kappa(\mathcal{T}_\mathcal{E}-\mathcal{E}_{u}) \cdot \kappa(\mu_{u}-\mathcal{T}_{\mu}) \cdot (1/(1+\beta \cdot \omega_{l})).
    \label{eq: loss_weights}
\end{equation}
Here, $\alpha$ and $\beta$ are user-defined scaling factors. $\kappa$ denotes the ReLU function, where $\kappa(x) = x$ if $x>0$ and $0$ otherwise. $\mathcal{T}_\mathcal{E}$ and $\mathcal{T}_{\mu}$ are learnable thresholds. $\mathcal{E}_{u} = \left\|(I_l - I) \odot y_{u}\right\|_{1}$ is the mean $\ell_{1}$ loss over the $u$-th region, where $\odot$ denotes the element-wise product. $\mu_{u}$ is the fraction of region $u$ that belongs to the hand foreground, computed as $\mu_{u} = \left\|y_{u} \odot y_{h}\right\|_{0}/\left\|y_{h}\right\|_{0}$. Therefore, Eq.~(\ref{eq: loss_weights}) can be viewed as a piecewise function that selects hand regions (via $\kappa(\mu_{u}-\mathcal{T}_{\mu})$) with acceptable rendering quality (via $\kappa(\mathcal{T}_\mathcal{E}-\mathcal{E}_{u})$). Additionally, we incorporate the temporal weight $\omega_l$ of the given frame into Eq.~(\ref{eq: loss_weights}) to modulate $\lambda_{u}$.

Consequently, the overall weighted mask that combines all regions is defined as
\begin{equation}
        \mathcal{W} =\sum_{u=1}^{U}\lambda_u \cdot y_u + (1-y_h) \cdot \kappa(\mathcal{T}_{b} - \left\|I_l-I\right\|_{1}),
    \label{eq: loss_weights_all}
\end{equation}
where the second term in Eq.~(\ref{eq: loss_weights_all}) controls the weight of the background, and $\mathcal{T}_{b}$ is also a learnable threshold. Fig.~\ref{fig: loss_weight} shows a scene with hand occlusions, where the proposed PAO strategy assigns higher weights to visible hand regions.

\subsection{Optimization Objectives}
We employ the weighted mask $\mathcal{W}$ computed by the PAO strategy to optimize WildGHand by minimizing the following loss function:
\begin{equation}
    \mathcal{L} = \mathcal{W} \odot \mathcal{L}_{\text{rec}}(I_l, I) + \mathcal{L}_{\text{reg}},
    \label{eq:loss}
\end{equation}
where $\mathcal{L}_{\text{rec}}$ denotes the reconstruction loss \cite{huang2024learning} between the input frame $I_l$ and the rendered image $I$. $\mathcal{L}_{\text{reg}}$ is a regularization term consisting of the following components:
\begin{equation}
    \begin{aligned}
    \mathcal{L}_{\text{reg}}
    = \sum_{n=1}^{N}
      & \lambda_{\text{bias}} \|\Delta g_n\|_{1}
      + \lambda_{\xi} \|\xi_n - 1\|_{2}
      + \lambda_{o} \|o_n - 1\|_{2} \\
      & +\, \lambda_{\text{Lap}} \mathcal{L}_{\text{Lap}}
      - \lambda_{\mathcal{W}} \|\mathcal{W}\|_{1}.
      \label{eq: loss_reg}
    \end{aligned}
\end{equation}

Here, $\lambda_{\text{bias}}$, $\lambda_{\xi}$, $\lambda_{o}$, $\lambda_{\text{Lap}}$, and $\lambda_{\mathcal{W}}$ are user-defined loss weights. The first three terms of Eq.~(\ref{eq: loss_reg}) impose point-wise regularization on the Gaussian attribute bias $\Delta g$, the shadow coefficient $\xi$, and the opacity $o$, respectively. $\mathcal{L}_{\text{Lap}}$ is a Laplacian regularizer \cite{moon2024expressive} that improves the connectivity between vertices. Finally, we incorporate the term $\left\|\mathcal{W}\right\|_{1}$ into the objective to prevent trivial weight assignments (e.g., all region weights being set to zero). With these weighted objectives, we can dynamically adjust the contribution of each region during optimization and thereby ensure accurate reconstruction of the input video.

\begin{table*}[!t]
  \caption{Detailed description of the proposed HWP dataset, covering hand type, primary action, recording device, number of frames, and the presence of perturbations.}
  \centering
  \small
  \setlength{\tabcolsep}{2pt}
  \renewcommand{\arraystretch}{1}
  \begin{tabular}{c|cccc|cccc}
    \toprule
    \multirow[c]{2}{*}{Capture} & \multirow[c]{2}{*}{Hand} & \multirow[c]{2}{*}{Action} & \multirow[c]{2}{*}{Capture Device} & \multirow[c]{2}{*}{$\#$Frames} &  \multicolumn{4}{c}{Perturbations}\\
    \cline{6-9}
    & & & & &Interaction & Complex Poses & Illumination Changes & Motion Blur \\
    \midrule
    0 & Single & Grabbing blocks  & Mobile phone & 901 & \checkmark  & \checkmark & & \\
    1 & Single & Multiple gestures  & Mobile phone & 742 & & \checkmark & & \checkmark \\
    2 & Single & Multiple gestures & Mobile phone & 828 & & &\checkmark & \\
    3 & Single & Spinning a pen & Mobile phone & 1086 & \checkmark & & & \checkmark \\
    4 & Interacting & Spinning a pen & Webcam & 1662 & \checkmark & & & \checkmark \\
    5 & Interacting & Sticking stickers & Webcam & 1216 & \checkmark & & & \checkmark \\
    6 & Interacting & Applying lotion & Webcam & 1524 & \checkmark & \checkmark & &  \\
    7 & Interacting & Multiple gestures & Webcam & 1360 &  & \checkmark & \checkmark &  \\
    8 & Interacting & Putting on a watch & Webcam & 1444 & \checkmark &\checkmark &  & \checkmark \\
    9 & Interacting & Scrolling on a phone & Mobile phone & 1533 & \checkmark &\checkmark & & \checkmark \\
    10 & Interacting & Shuffling cards & Mobile phone & 1525 & \checkmark & \checkmark & & \checkmark \\
    \bottomrule
  \end{tabular}
  \label{tab:dataset_details}
\end{table*}

During training on interacting-hand sequences, we introduce an additional $\ell_{1}$ regularization term that enforces consistency between the texture features of the left and right hands:
\begin{equation}
\mathcal{L}_{\text{cross}} = \|\mathbf{T}_{\text{left}} - \mathbf{T}_{\text{right}}\|_{1},
\end{equation}
where $\mathbf{T}_{\text{left}}$ and $\mathbf{T}_{\text{right}}$ denote the extracted texture features of the left and right hands, respectively.  
This term leverages the inherent bilateral symmetry of human hands, allowing visible regions of one hand to provide informative cues for reconstructing occluded regions on the other. Consequently, even without explicit joint modeling, the system can robustly infer missing textures and maintain appearance consistency under inter-hand occlusions. 

For clarity, we primarily focus on the single-hand setting in the main formulation, and the cross-hand consistency term is included only during training on interacting-hand sequences.

\section{HWP Dataset}
\begin{figure}[!t]
  \centering
\includegraphics[width=1.0\hsize]{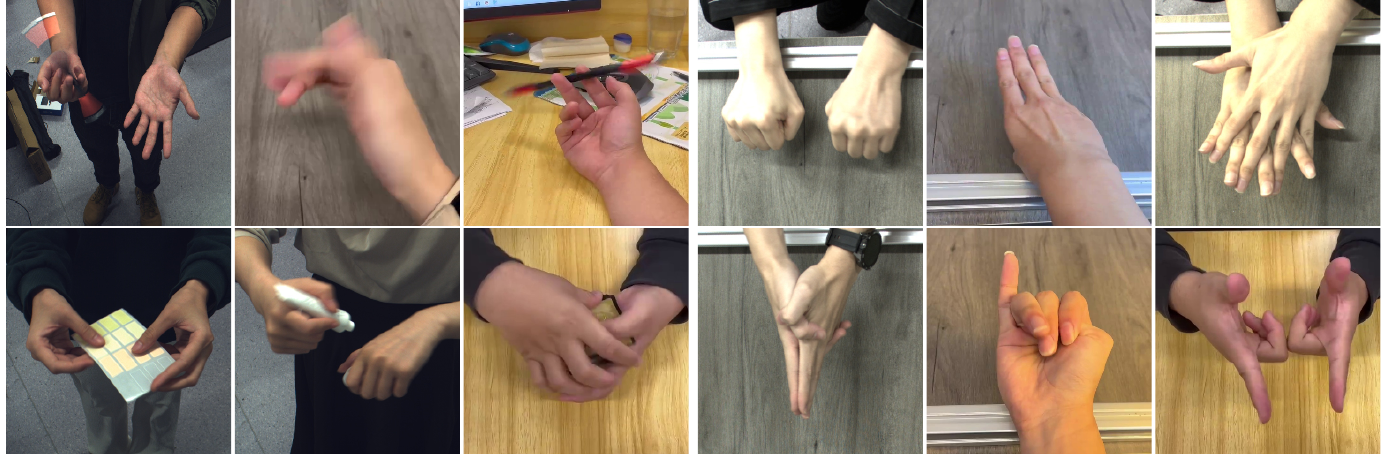}
   \caption{Visual examples of our HWP dataset, which covers diverse challenging scenes like motion blur, hand-object interactions, illumination variations, and complex poses (from top left to bottom right).}
   \label{fig:dataset_example}
\end{figure}

As shown in Table~\ref{tab:dataset_comparison}, most existing hand video datasets exhibit notable limitations, such as restricted interaction diversity, inadequate coverage of real-world perturbations, or the absence of clean and reliable test subsets. Therefore, we introduce the HWP dataset, a monocular hand video dataset captured in unconstrained environments to overcome these issues. 

Specifically, videos on HWP are recorded with either a fixed single camera or a handheld mobile phone. We target four dominant perturbation types: (i) hand–object interaction, (ii) complex poses, (iii) illumination change, and (iv) motion blur. To increase realism and interaction diversity, we consider daily actions and gestures such as spinning a pen or applying hand cream. In total, our dataset contains four single-hand and seven interacting-hand sequences, with over $13.8$K frames.

Table~\ref{tab:dataset_details} summarizes the composition of the proposed HWP dataset, including hand type (single or interacting), primary action category, recording device (fixed camera or handheld mobile phone), number of frames, and associated perturbation types. Representative visual samples from different sequences are shown in Figure~\ref{fig:dataset_example}.

In this way, the proposed HWP dataset encompasses a broad spectrum of systematically categorized perturbations under clearly categorized conditions while maintaining an isolated, clean test subset for objective benchmarking. This design enables both \textit{fine-grained evaluation under specific challenges} and \textit{fair cross-method comparison in realistic, in-the-wild scenarios}.

\section{Experiments}
\label{sec:Experiments}

In this section, we validate the effectiveness of the proposed WildGHand framework through extensive experiments. We conduct evaluations on publicly available datasets as well as self-collected and online videos that cover a wide range of in-the-wild perturbations.

\begin{table*}[!t]
  \centering
  \caption{Comparison with state-of-the-art methods on the three dataset, including WildGHand, InterHand2.6M and AnchorCrafter. ``Single'' and ``Inter.'' denote single-hand and interacting-hand scenarios, respectively.}
  % 调整字体与列间距
  \setlength{\tabcolsep}{3.3pt}  % ← 控制列间距（默认约6pt，可调至2.5–3.5）
  \renewcommand{\arraystretch}{0.95}  % ← 控制行距（默认1.0，可调至0.9–1.0）
  \small
  \begin{tabular}{lccc|ccc|ccc|ccc}
    \toprule
    \multirow{2}{*}{Method} & 
    \multicolumn{3}{c|}{\textbf{WildGHand (Single)}} & 
    \multicolumn{3}{c|}{\textbf{WildGHand (Inter.)}} & 
    \multicolumn{3}{c|}{\textbf{InterHand2.6M (Inter.)}} & 
    \multicolumn{3}{c}{\textbf{AnchorCrafter (Inter.)}} \\
    \cmidrule(lr){2-13}
     & PSNR$\uparrow$ & SSIM$\uparrow$ & LPIPS$\downarrow$ 
     & PSNR$\uparrow$ & SSIM$\uparrow$ & LPIPS$\downarrow$
     & PSNR$\uparrow$ & SSIM$\uparrow$ & LPIPS$\downarrow$
     & PSNR$\uparrow$ & SSIM$\uparrow$ & LPIPS$\downarrow$ \\
    \midrule
    UHM~\cite{moon2024authentic} & 20.24 & 0.867 & 0.203 & -- & -- & -- & -- & -- & -- & -- & -- & -- \\
    Handy~\cite{potamias2023handy} & 22.99 & 0.874 & 0.127 & 24.10 & 0.864 & 0.141 & 24.34 & 0.867 & 0.198  &25.88  & 0.912  & 0.074 \\
    InterGaussianHand~\cite{huang2024learning} & 23.06 & 0.858 & 0.147 & 24.69 & 0.851 & 0.148 & 26.70 & 0.861 & 0.173 & 24.28  & 0.912  & 0.098 \\
    \textbf{WildGHand (Ours)} & \textbf{26.24} & \textbf{0.916} & \textbf{0.116} & \textbf{26.29} & \textbf{0.893} & \textbf{0.122} &\textbf{28.25}   & \textbf{0.894}   & \textbf{0.151}   &\textbf{27.68} & \textbf{0.948} & \textbf{0.072} \\
    \bottomrule
  \end{tabular}
  \label{tab:comparison}
\end{table*}

\begin{table}[t]
\centering
\caption{Comparison on online videos.}
\label{tab:comp_online}
\small
\begin{tabular}{lccc}
\toprule
Method & PSNR$\uparrow$ & SSIM$\uparrow$ & LPIPS$\downarrow$ \\
\midrule
    Handy~\cite{potamias2023handy} &21.46  & 0.809  & 0.191 \\
    InterGaussianHand~\cite{huang2024learning} & 22.95  & 0.820  & 0.157 \\
    \textbf{WildGHand (Ours)} &\textbf{27.39} & \textbf{0.920} & \textbf{0.080} \\
\bottomrule
\end{tabular}
\end{table}

\subsection{Datasets}
We evaluate the proposed method on our HWP dataset, InterHand2.6M~\cite{moon2020interhand2}, the AnchorCrafter dataset~\cite{xu2024AnchorCrafter}, and in-the-wild online videos. For each video, we use $80\%$ of the frames for optimization, $10\%$ for validation, and $10\%$ for testing. Details of the three datasets other than HWP are summarized below:

 \textbf{InterHand2.6M.}
InterHand2.6M~\cite{moon2020interhand2} is a widely used in-lab benchmark for hand--hand interactions. We evaluate on three sequences: \texttt{Capture21/0390\_dh\_touchROM/cam400015}, \texttt{Capture21/0390\_dh\_touchROM/cam400016}, and \texttt{Capture21/0286\_handscratch/cam400016}.

 \textbf{AnchorCrafter dataset.}
The AnchorCrafter dataset~\cite{xu2024AnchorCrafter} provides in-the-wild human--object interaction videos. We use the sequence \texttt{tune/1\_0} to assess performance on hand–object interaction scenarios.

 \textbf{Online videos.}  
To further evaluate generalization, we test on hand videos collected from online platforms, which feature diverse lighting conditions, camera motions, backgrounds, and occlusions.

\subsection{Setup}
\label{sec:Setup}

\textbf{Implementation Details}. Following \cite{huang2024learning}, we pretrain the Gaussian attribute prediction network on the large-scale InterHand2.6M dataset \cite{moon2020interhand2} (licensed under CC BY-NC 4.0). For each video, we optimize our model on two NVIDIA A6000 GPUs with a batch size of $4$ for $50$ epochs using the Adam optimizer \cite{kingma2014adam}. The initial learning rate is set to $1 \times 10^{-6}$ for the learnable thresholds and to $1 \times 10^{-4}$ for the remaining parameters; both rates are decayed by a factor of $0.5$ every five epochs. The feature dimension of the latent identity map $D$ is set to $33$. The loss weights in Eq.~(\ref{eq: loss_reg}) are set to $\lambda_{\text{bias}} = 0.1$, $\lambda_{\xi} = 0.001$, $\lambda_{o} = 0.1$, $\lambda_{\text{Lap}} = 0.1$, and $\lambda_{\mathcal{W}} = 0.5$. The initial learnable thresholds for interacting-hand videos are set to $\mathcal{T}_\mathcal{E}=1.0$, $\mathcal{T}_{\mu}=0.3$, and $\mathcal{T}_{b}=10.0$, while those for single-hand videos are set to $\mathcal{T}_\mathcal{E}=1.0$, $\mathcal{T}_{\mu}=0.3$, and $\mathcal{T}_{b}=3.0$.

\textbf{Baselines and Metrics}.
We compare WildGHand with three pioneering hand avatar methods, including UHM \cite{moon2024authentic}, Handy \cite{potamias2023handy}, and InterGaussianHand \cite{huang2024learning}. All methods are trained and evaluated under the same experimental setting based on their open-source implementations. Among these methods, both UHM and Handy adopt their own hand models. However, UHM only provides its right-hand model fitted on its own dataset, which is not compatible with MANO-HD and hard to convert. Hence, we report only the single-hand performance for UHM. As for Handy, it is compatible with MANO-HD, and hence we convert it accordingly and optimize the latent vectors of its texture model using differentiable rendering. Following \cite{potamias2023handy, karunratanakul2023harp, moon2024authentic, huang2024learning}, we evaluate the quality of rendered images using PSNR \cite{sara2019image}, LPIPS \cite{zhang2018unreasonable}, and SSIM \cite{wang2004image}.

\begin{figure*}[!t]
  \centering
   \includegraphics[width=1.0\hsize]{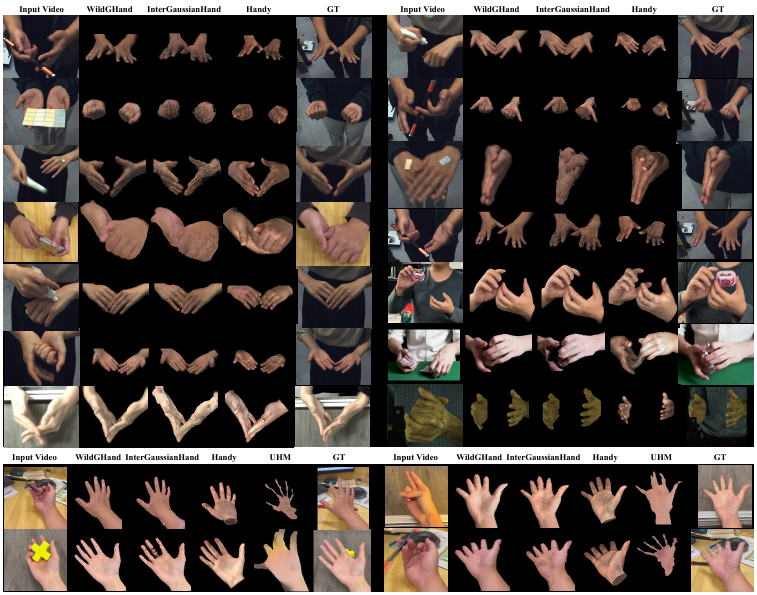}
   \caption{Qualitative comparisons between our proposed WildGHand model with state-of-the-art methods on interacting-hand videos (top) and single-hand videos (bottom).}
   \label{fig: comparison}
\end{figure*}

\begin{figure*}[!t]
  \centering
   \includegraphics[width=1.0\hsize]{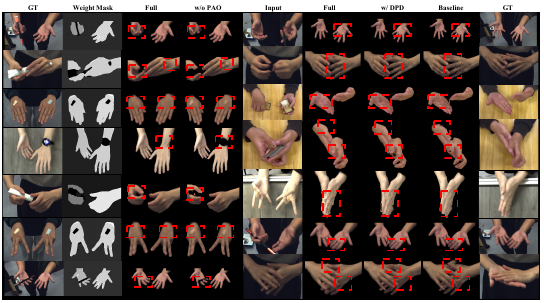}
   \caption{Visual example of the ablation study on the proposed modules. Left: Weight masks learned by the proposed PAO strategy, light hand regions indicate higher weights. Right: Results of the variants of the proposed method. The full model obtains the best results consistently on various videos.}
   \label{fig: ablation}
\end{figure*}

\subsection{Comparison with State-of-the-art Methods}

\begin{table*}[!t]
  \caption{Comparison with state-of-the-art methods on the interacting-hand videos of the HWP dataset, evaluated across four scenarios: hand-object interactions, complex poses, illumination variations, and motion blur.}
  \centering
  \setlength{\tabcolsep}{4.5pt}
  \renewcommand{\arraystretch}{1.05}
  \begin{tabular}{lccc|ccc|ccc|ccc}
    \toprule
    \multirow[c]{2}{*}{Method} & \multicolumn{3}{c|}{\textbf{Hand-object Interactions}} & \multicolumn{3}{c|}{\textbf{Complex Poses}} & \multicolumn{3}{c|}{\textbf{Illumination Variations}} & \multicolumn{3}{c}{\textbf{Motion Blur}} \\
    \cline{2-13}
     & PSNR↑ & SSIM↑ & LPIPS↓ & PSNR↑ & SSIM↑ & LPIPS↓ & PSNR↑ & SSIM↑ & LPIPS↓ & PSNR↑ & SSIM↑ & LPIPS↓ \\
    \midrule
     Handy & 24.67 & 0.872 & 0.138 & 24.14 & 0.863 & 0.138 & 20.65 & 0.814 & 0.156 & 24.33 & 0.872 & 0.139 \\
    InterGaussianHand & 25.11 & 0.858 & 0.146 & 24.51 & 0.848 & 0.142 & 22.18 & 0.805 & 0.158 & 24.71 & 0.860 & 0.145 \\
    \textbf{WildGHand (ours)} & \textbf{26.87} & \textbf{0.902} & \textbf{0.118} & \textbf{25.29} & \textbf{0.881} & \textbf{0.124} & \textbf{22.81} & \textbf{0.839} & \textbf{0.148} & \textbf{26.68} & \textbf{0.901} & \textbf{0.116} \\
    \bottomrule
  \end{tabular}
  \label{tab:comparison_dis}
\end{table*}

\textbf{Quantitative Comparison}.
Table~\ref{tab:comparison} reports quantitative results of our method and state-of-the-art approaches on HWP, InterHand2.6M, and AnchorCrafter. The proposed WildGHand framework consistently outperforms all baselines across all metrics in both single-hand and interacting-hand settings, demonstrating its effectiveness and robustness. Table~\ref{tab:comp_online} further reports results on Internet videos, highlighting the strong generalization ability of our method under real-world conditions.

Our previous InterGaussianHand is designed for one-shot reconstruction and relies on well-captured images and accurate hand mesh estimation. As a result, it is highly susceptible to perturbations in target frames, which leads to a substantial performance drop. Similarly, Handy integrates explicit hand meshes with implicit texture optimization and is therefore sensitive to perturbations; moreover, its texture optimization tends to overfit to irrelevant details, further degrading performance. Because UHM is incompatible with MANO-HD, it relies on off-the-shelf estimators to obtain foreground masks and joint coordinates and then optimizes its animation objective. However, this process is often compromised by perturbations, resulting in unstable fitting and degraded performance. In contrast, by incorporating the DPD module and the PAO strategy, WildGHand effectively mitigates these issues and produces high-quality results.

\textbf{Qualitative Comparison}.
Fig.~\ref{fig: comparison} shows qualitative comparisons between our approach and the baseline methods on both interacting-hand and single-hand videos. InterGaussianHand fails to generate accurate hand geometry and appearance in the presence of perturbations. Without a dedicated mechanism for handling interference and occlusions, it is heavily influenced by moving objects and motion blur, leading to suboptimal results. Handy, while capable of recovering basic hand geometry and a reasonable appearance, fails to capture fine-grained identity cues. Although its texture optimization in the pretrained latent space is relatively stable, it is still misled by perturbations, compromising the final results. For instance, the skin tones synthesized by Handy deviate significantly from the ground truth, particularly in interacting-hand scenarios. UHM is affected by unreliable foreground masks and joint coordinates, producing distorted and unnatural hand structures (e.g., the first row of the single-hand results). In contrast, WildGHand successfully reconstructs intricate hand details, such as nails, wrinkles, and veins, across diverse views and poses. These qualitative comparisons further validate the robustness and strong performance of our approach in generating accurate and realistic hand avatars.

\subsection{Ablation Study}

We conduct ablation studies on our dataset to validate the effectiveness of the proposed DPD module and the PAO strategy. We consider the Gaussian attribute prediction network as the baseline.

\begin{table}[!t]
  \caption{Ablation study on the proposed modules.}
  \centering
  \small
  % 调整字体与列间距
  \setlength{\tabcolsep}{3.0pt}  % ← 控制列间距（默认约6pt，可调至2.5–3.5）
  \begin{tabular}{ccccc|ccc}
    \toprule
     \multirow[c]{2}{*}{DPD} & \multirow[c]{2}{*}{PAO} & \multicolumn{3}{c|}{\textbf{Single Hand}} & \multicolumn{3}{c}{\textbf{Interacting Hands}} \\
    \cmidrule(lr){3-8}
     & & PSNR↑    & SSIM↑    & LPIPS↓  & PSNR↑    & SSIM↑    & LPIPS↓ \\
    \midrule
      &  & 22.65 & 0.862 & 0.151  &24.86  & 0.867 & 0.123 \\
      \checkmark & & 23.00  & 0.869 & 0.138  &25.01 & 0.868 & 0.122 \\
      \checkmark& \checkmark & \textbf{26.24} & \textbf{0.916} & \textbf{0.116}  &\textbf{26.29} & \textbf{0.893} & \textbf{0.122}\\
    \bottomrule
  \end{tabular}
  \label{tab: ablation}
\end{table}

\textbf{Effectiveness of DPD}. Our lightweight DPD module has $75k$ learnable parameters, roughly $0.018\%$ of the whole model ($400$M). Despite its tiny capacity, DPD still boosts the performance of the baseline (Table \ref{tab: ablation}) and addresses a few artifacts like floaters and broken structures (Figure \ref{fig: ablation}).

\textbf{Effectiveness of PAO}. As reported in Table \ref{tab: ablation}, the performance gains achieved with PAO are more substantial than those by DPD. This is reasonable, as merely increasing network capacity without a proper optimization goal does not necessarily guarantee enhanced robustness against perturbations. Figure \ref{fig: ablation} also shows the weighted masks predicted by PAO, which clearly indicate the areas of artifacts generated by the baseline. Guided by these weighted masks, the full WildGHand model avoids overfitting to perturbations and successfully eliminates artifacts in the rendered images. We also notice that the performance gains in single-hand scenarios are significantly greater than those in interacting-hand cases (e.g., a $23.1\%$ relative reduction in the LPIPS score is observed in the former). This is caused by the increased difficulty of preserving hand information in single-hand videos when perturbations occur, which the proposed method manages to tackle. 

\textbf{Robustness to perturbation}. Table \ref{tab:comparison_dis} compares WildGHand with two state-of-the-art methods, Handy and InterGaussianHand, under four challenging scenarios: hand-object interactions, complex poses, illumination variations, and motion blur. WildGHand consistently outperforms both baselines, achieving the best PSNR, SSIM, and lowest LPIPS across all settings. These results demonstrate WildGHand’s robustness and effectiveness across diverse real-world perturbations.

\textbf{Binary mask vs.\ PAO}.
We compare our perturbation-aware occlusion (PAO) strategy with a SAM-based binary mask. 
As shown in Table~\ref{tab:pao_vs_binary}, binary masks lack geometric constraints and fail to reliably distinguish occluders from the background under challenging in-the-wild conditions. In contrast, PAO adaptively down-weights unreliable regions and preserves informative hand areas, resulting in improved reconstruction quality.

\begin{table}[t]
\centering
\caption{Comparison between PAO and SAM-based binary mask.}
\label{tab:pao_vs_binary}
\small
\begin{tabular}{lccc}
\toprule
Method & PSNR$\uparrow$ & SSIM$\uparrow$ & LPIPS$\downarrow$ \\
\midrule
\textbf{WildGHand (ours)} & \textbf{26.44} & \textbf{0.871} & \textbf{0.179} \\
WildGHand w/ binary mask & 25.36 & 0.844 & 0.184 \\
\bottomrule
\end{tabular}
\end{table}

\begin{table}[t]
\centering
\caption{Ablation on prediction heads.}
\label{tab:head_ablation}
\small
\begin{tabular}{lccc}
\toprule
Method & PSNR$\uparrow$ & SSIM$\uparrow$ & LPIPS$\downarrow$ \\
\midrule
\textbf{WildGHand (ours)} & \textbf{26.44} & \textbf{0.871} & \textbf{0.179} \\
w/o Heads (iii) \& (iv)  & 26.20 & 0.864 & 0.177 \\
\bottomrule
\end{tabular}
\end{table}

\textbf{Effectiveness of prediction heads}.
Our network employs four prediction heads (Sec.~\ref{sec: framework}):  
(i)–(ii) estimate the core Gaussian attributes (e.g., color, position, shading), which are standard in 3D Gaussian Splatting models;  
(iii) predicts vertex offsets to capture shape and pose variations; and  
(iv) refines hand-specific parameters via residual correction.  
As shown in Table~\ref{tab:head_ablation}, removing heads (iii) and (iv) leads to measurable degradation in both reconstruction accuracy and perceptual quality.

\subsection{Bias strength and interpretation.}
As stated in Sec.~\ref{sec: DPD}, our biases model frame-specific perturbations. Consequently, larger bias magnitudes imply stronger perturbations and are not favored. This can be validated by Figure~\ref{fig: loss_weight}, where temporal weights correlate with visual disturbance. We further report average temporal weights across captures of different noise levels in Table \ref{tab:bias_strength}. It is clear that higher perturbation levels lead to larger weights and lower image quality, confirming our model’s adaptive behavior.

\begin{table}[t]
\centering
\caption{Average temporal weight vs.\ perturbation level. 
Higher perturbation correlates with larger weights and degraded quality. 
Weights are scaled by $10^{-3}$ for readability.}
\label{tab:bias_strength}
\fontsize{7.2pt}{8.6pt}\selectfont
\setlength{\tabcolsep}{3.3pt}  % ← 控制列间距（默认约6pt，可调至2.5–3.5
\begin{tabular}{cccccc}
\toprule
Capture & Disturbance & PSNR $\uparrow$ & SSIM $\uparrow$ & LPIPS $\downarrow$  & Avg.\ T.\ Wt.\,$(\times 10^{-3})$ $\uparrow$ \\
\midrule
4 & Low    & 29.59 & 0.931 & 0.115 & 0.234 \\
5 & Medium & 28.00 & 0.915 & 0.116 & 1.581 \\
6 & High   & 27.83 & 0.907 & 0.123 & 7.572 \\
\bottomrule
\end{tabular}
\end{table}

\begin{table}[!t]
    \caption{Mean and standard deviation of the performance of InterGaussianHand and WildGHand.}
    \label{tab:standard_deviation}
    \centering
    \scriptsize
    \begin{tabular}{lccc}
        \toprule
        Method     & PSNR↑    & SSIM↑    & LPIPS↓     \\
        \midrule
        InterGaussianHand     &26.2518$\pm$0.0770   & 0.8723$\pm$0.0009   & 0.1582$\pm$0.0012  \\
        \textbf{WildGHand (ours)}     &\textbf{29.5859$\pm$0.1028}   & \textbf{0.9313$\pm$0.0015}   & \textbf{0.1146$\pm$0.0029}  \\
        \bottomrule
    \end{tabular}
\end{table}

\section{Conclusion}
\label{sec:Conclusion}
In this paper, we present WildGHand for reconstructing high-fidelity hand avatars from monocular in-the-wild videos under diverse perturbations. WildGHand introduces a dynamic perturbation disentanglement module that models perturbations as temporally weighted biases of 3D Gaussian attributes during optimization, and removes these biases at inference to mitigate the risk of overfitting to perturbations. Furthermore, WildGHand introduces a perturbation-aware optimization strategy that generates anisotropic weighted masks to unreliable regions and provides robust supervision. We evaluate WildGHand on two public datasets and our collected dataset covering both single-hand and interacting-hand scenarios. Results demonstrate that WildGHand consistently produces high-quality hand avatars and outperforms state-of-the-art methods across multiple qualitative and quantitative metrics.

\bibliographystyle{IEEEtran}
\bibliography{main}

\begin{IEEEbiography}
[{\includegraphics[width=1in,height=1.25in,clip,keepaspectratio]{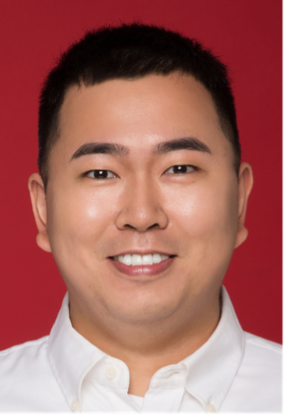}}]
{Hanhui Li}
received the Ph.D. degree in computer software and theory from Sun Yat-sen University, Guangzhou, China, in 2018, where he also received the B.S. degree in computer science and technology in 2012. He is now a research associate professor in Shenzhen Campus of Sun Yat-sen University, China. Before that, he was a research fellow in Nanyang Technological University, Singapore, from 2019 to 2021. His research interests include image processing, 3D content creation, and artificial intelligence generated content.
\end{IEEEbiography}

\vspace{-1cm}

\begin{IEEEbiography}
[{\includegraphics[width=1in,height=1.25in,clip,keepaspectratio]{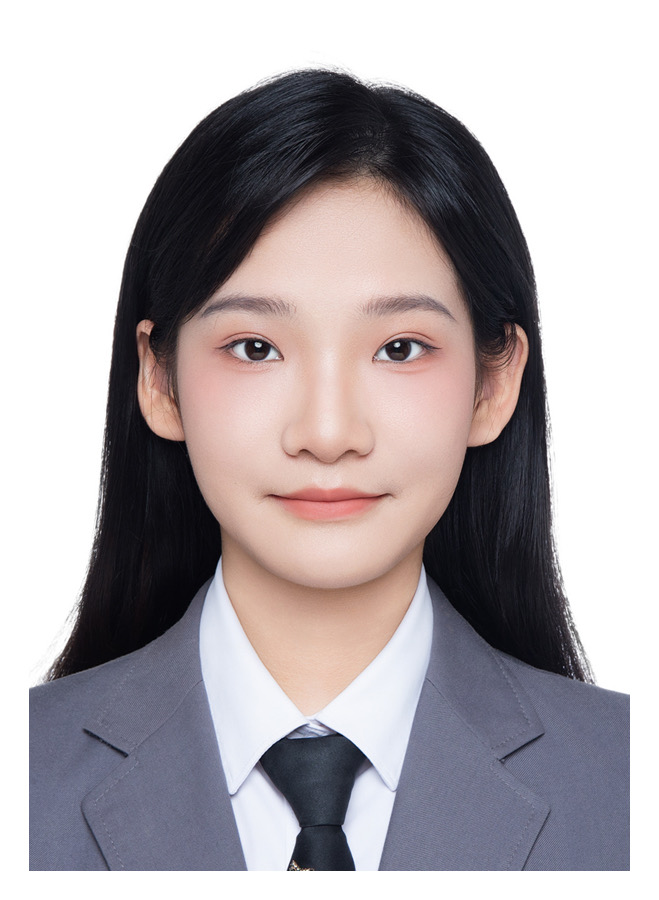}}]
{Xuan Huang}
received the B.E. degree in Intelligent Engineering from the School of Intelligent Systems Engineering, Sun Yat-sen University, Shenzhen, Guangdong, China, in 2023. She is currently pursuing the M.E. degree in Intelligent Engineering at the same institution. She has published papers in top-tier conferences, including NeurIPS and AAAI. Her research interests focus on 3D reconstruction, video generation, and their applications in digital humans.
\end{IEEEbiography}

\vspace{-1cm}

\begin{IEEEbiography}
[{\includegraphics[width=1in,height=1.25in,clip,keepaspectratio]{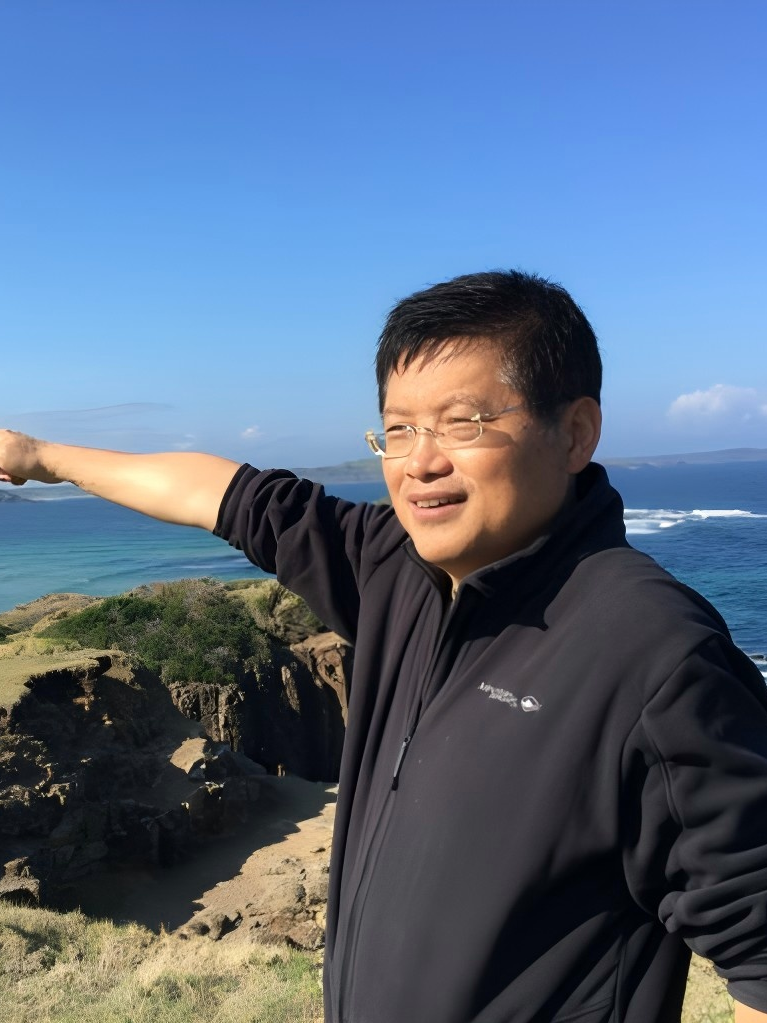}}]
{Wanquan Liu}
received the B.S. degree in Applied Mathematics from Qufu Normal University, China, in 1985, the M.S. degree in Control Theory and Operation Research from Chinese Academy of Science in 1988, and the Ph.D. degree in Electrical Engineering from Shanghai Jiaotong University, in 1993. He once held ARC, U2000 and JSPS Fellowships and secured over \$2.4 million in research funds from various sources. He is a Full Professor at the School of Intelligent Systems Engineering, SYSU, Guangzhou, China. His research interests include large-scale pattern recognition, machine learning, and control systems.
\end{IEEEbiography}

\vspace{-1cm}

\begin{IEEEbiography}
[{\includegraphics[width=1in,height=1.25in,clip,keepaspectratio]{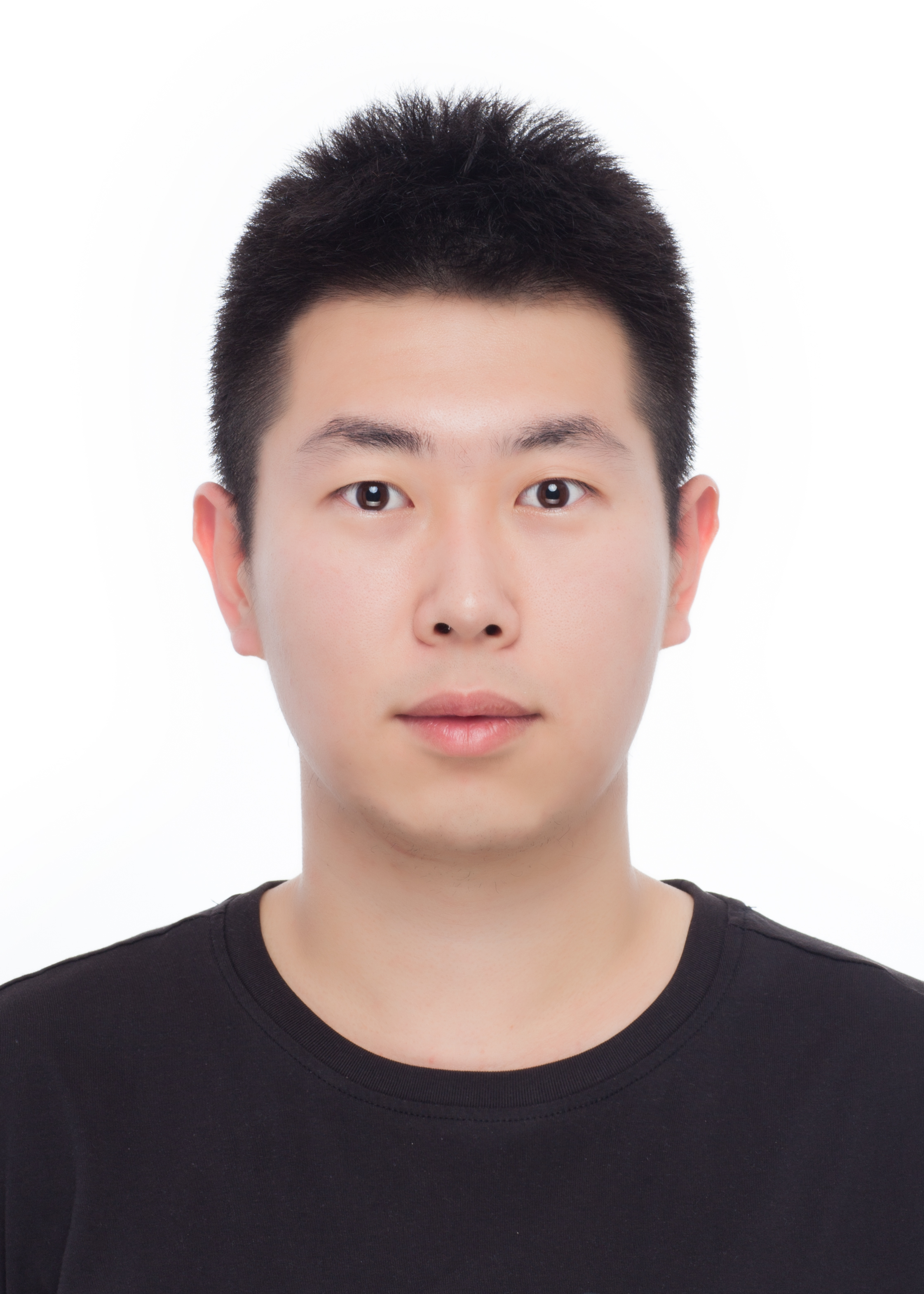}}]
{Yuhao Cheng}
received the B.E. degree from the Beijing University of Posts and Telecommunications, Beijing, China, in 2018, the First-Class Honor B.E. degree from Queen Mary University of London, London, U.K. in 2018, and the Master’s degree from Beijing University of Posts and Telecommunications, Beijing, China, in 2021. He is currently working in Lenovo Research. His current research interests include the Large Language Model, human-centric computer vision, video representation, and so on.
\end{IEEEbiography}

\vspace{-1cm}

\begin{IEEEbiography}
[{\includegraphics[width=1in,height=1.25in,clip,keepaspectratio]{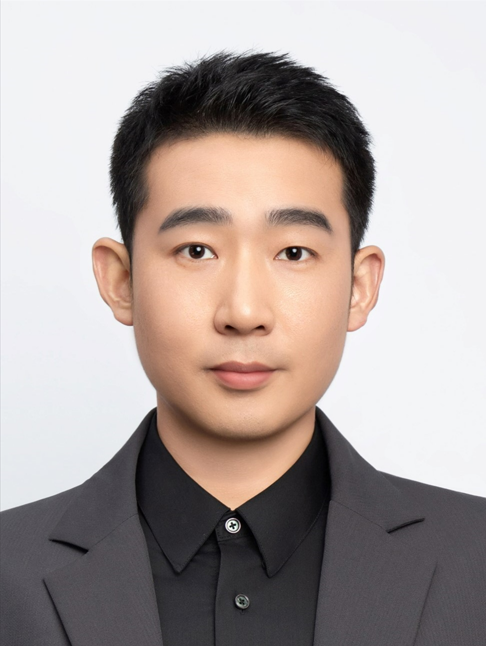}}]
{Long Chen}
received his B.S. degree from Northeastern University, Shenyang, China, in 2016, and the Master's degree from Tianjin University, Tianjin, China, in 2019. He is currently the advisory researcher in Lenovo Research. His research interests include computer vision, image generation, and pattern recognition.
\end{IEEEbiography}

\vspace{-1cm}

\begin{IEEEbiography}
[{\includegraphics[width=1in,height=1.25in,clip,keepaspectratio]{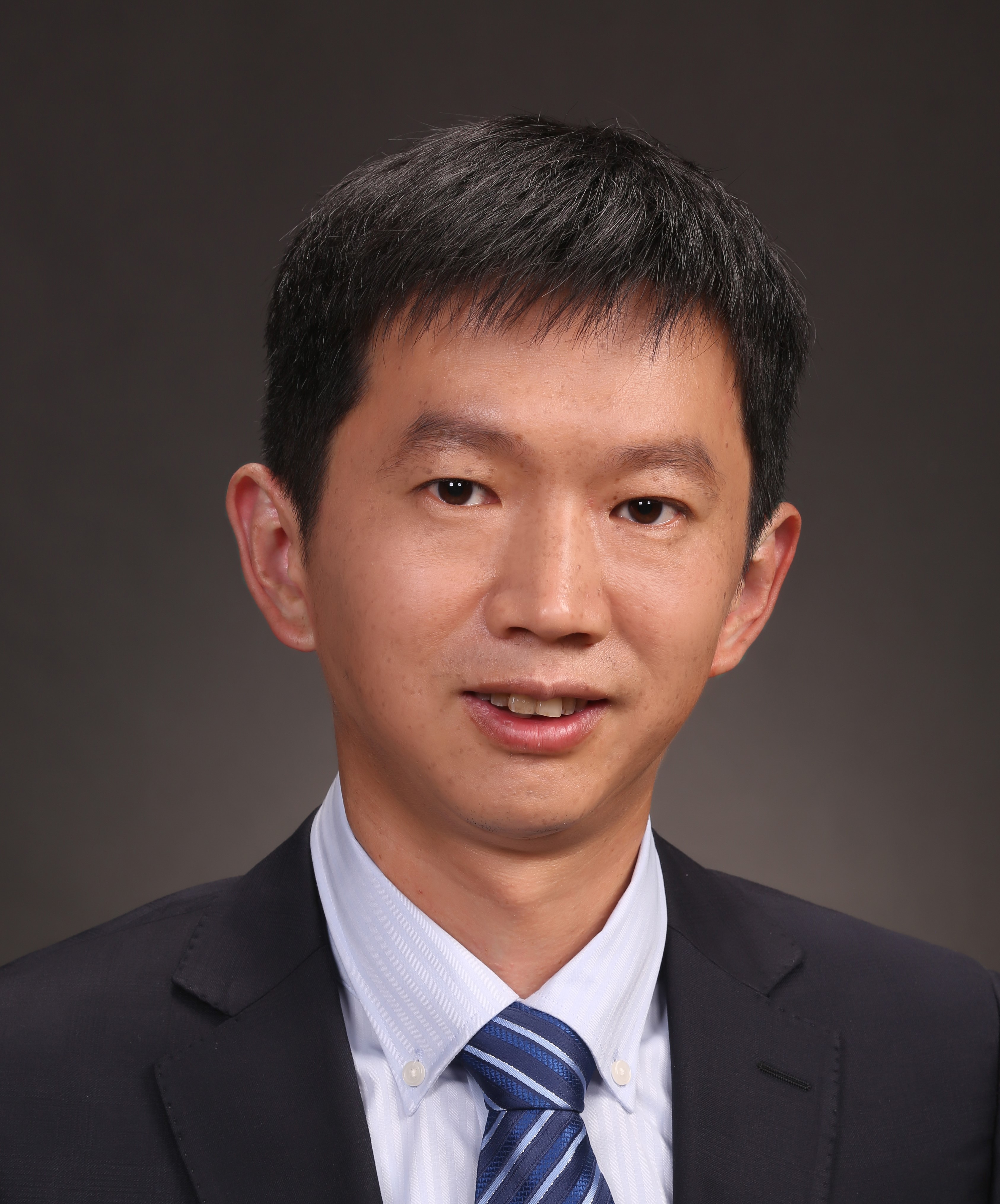}}]
{Yiqiang Yan}
is the Distinguished Researcher of Lenovo Group, and has been engaged in the development of smart devices for a long time. He has led the development of many industry-first devices, including wireless display devices, smart TVs, flexible devices, smart retail, and tablet/laptop two-in-one devices. These products he led to develop have won more than 120 awards at CES and MWC, and won the CCF Technology Invention Award in 2021. Currently, he leads the PC innovation of computing architecture and human-computer interaction for large language models. He has published 78 invention patents, including 19 U.S. patents.
\end{IEEEbiography}

\vspace{-1cm}

\begin{IEEEbiography}
[{\includegraphics[width=1in,height=1.25in,clip,keepaspectratio]{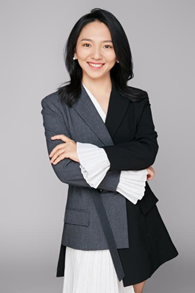}}]
{Xiaodan Liang}
 is currently a Professor at Sun Yat-sen University. She was a postdoc researcher in the machine learning department at Carnegie Mellon University, working with Prof. Eric Xing, from 2016 to 2018. She received her Ph.D. degree from Sun Yat-sen University in 2016. She has published several cutting-edge projects on human-related analysis, including human parsing, pedestrian detection, and instance segmentation, 2D/3D human pose estimation, and activity recognition.
\end{IEEEbiography}

\vspace{-1cm}

\begin{IEEEbiography}
[{\includegraphics[width=1in,height=1.25in,clip,keepaspectratio]{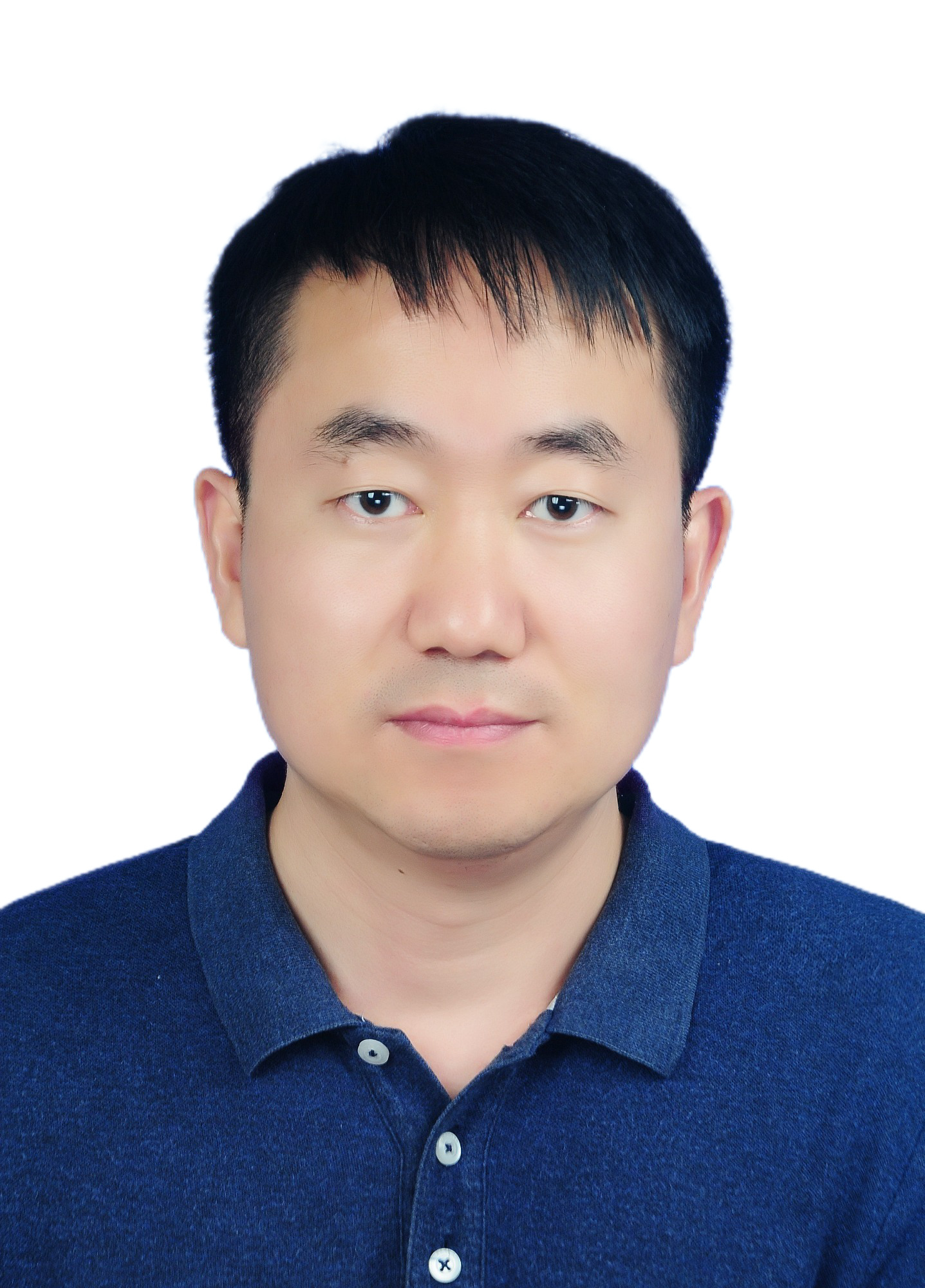}}]
{Chenqiang Gao}
 received the B.S. degree in computer science from the China University of Geosciences, Wuhan, China, in 2004 and the Ph.D. degree in control science and engineering from the Huazhong University of Science and Technology, Wuhan, China, in 2009. In August 2009, he joined the School of Communications and Information Engineering, Chongqing University of Posts and Telecommunications (CQUPT), Chongqing, China. In September 2012, he joined the Informedia Group with the School of Computer Science, Carnegie Mellon University, Pittsburgh, PA, USA, working on multimedia event detection (MED) and surveillance event detection (SED) until March 2014, when he returned to CQUPT. In September 2023, he joined the School of Intelligent Systems Engineering, Sun Yat-sen University, Shenzhen, Guangdong, China. His research interests include image processing, infrared target detection, action recognition, and event detection.
\end{IEEEbiography}

\end{document}